\documentclass[journal]{IEEEtran}
\usepackage{amsmath,amsfonts}
\usepackage{algorithmic}
\usepackage{algorithm}
\usepackage{array}
\usepackage{textcomp}
\usepackage{stfloats}
\usepackage{url}
\usepackage{verbatim}
\usepackage{graphicx}
\usepackage{amsmath}
\usepackage{multirow}
\usepackage{multicol}
\usepackage{hyperref}
\usepackage{amssymb}
\usepackage{booktabs}
\usepackage{color}
\usepackage{cite}
\usepackage{pifont}       
\usepackage{bbding}      
\usepackage{fontawesome} 
\usepackage{subcaption}
\hypersetup{
	colorlinks=true,
	linkcolor=blue,
	citecolor=green,
	urlcolor=magenta
}
\newcommand{\dd}{\,{\mathrm d}}

\begin{document}

\title{Space-Time Video Super-resolution \\ with Neural Operator}

\author{
Yuantong Zhang,~\IEEEmembership{Student~Member,~IEEE}, Hanyou Zheng, Daiqin Yang, Zhenzhong Chen,~\IEEEmembership{Senior~Member,~IEEE} 
Haichuan Ma, 
and Wenpeng Ding
\thanks{

 (Corresponding author: Zhenzhong Chen)  }
}

\maketitle

\begin{abstract}
This paper addresses the task of space-time video super-resolution (ST-VSR). Existing methods generally suffer from inaccurate motion estimation and motion compensation (MEMC) problems for large motions. Inspired by recent progress in physics-informed neural networks, we model the challenges of MEMC in ST-VSR  as a mapping between two continuous function spaces.
Specifically, our approach transforms independent low-resolution representations in the coarse-grained continuous function space into refined representations with enriched spatiotemporal details in the fine-grained continuous function space.
To achieve efficient and accurate MEMC, we design a Galerkin-type attention function to perform frame alignment and temporal interpolation. Due to the linear complexity of the Galerkin-type attention mechanism, our model avoids patch partitioning and offers global receptive fields, enabling precise estimation of large motions. The experimental results show that the proposed method surpasses state-of-the-art techniques in both fixed-size and continuous space-time video super-resolution tasks.
\end{abstract}


\section{Introduction}
\label{sec:intro}
The rapid advancement of multimedia technology, especially in hardware devices, has resulted in the widespread availability of high-resolution (HR) and high-frame-rate video sequences.
However, due to shooting conditions and storage device limitations, recorded videos are often stored with limited spatial resolution and frame rates. 
Space-time video super-resolution (ST-VSR) seeks to transform low-resolution and low-frame-rate videos to higher spatial and temporal resolutions simultaneously, finding broad applications across various domains \cite{flynn2016deepstereo,SuDWSHW17}. The independent execution of video frame interpolation (VFI) and video super-resolution (VSR) overlooks the inherent correlation between space and time. Therefore, recent research has shifted towards considering VFI and VSR as a unified space-time video super-resolution (ST-VSR) process. Some earlier methods \cite{2021Zooming,xu2021temporal} often employ ConvLSTM \cite{2017DeepRain} structures for propagating information across multiple frames and subsequently performing upsampling. More recently, to increase model adaptability, several continuous space-time video super-resolution methods \cite{shi2021learning,chen2022videoinr,chen2023motif} have been introduced. These methods allow input modulation from low frame rate and low resolution to produce arbitrary high frame rate and high-resolution outcomes.

Nevertheless, despite the considerable progress made by these methods, two critical challenges still need to be addressed:  \emph{ 1) efficient  MEMC}, \emph{2) the ability to handle extreme motion.} 
Regarding the first challenge, existing solutions either handle alignment and temporal interpolation separately \cite{2020Space,2021Zooming,xu2021temporal} for pre-existing frames and interpolated frames or introduce additional encoders \cite{chen2022videoinr,chen2023motif} or separate optical flow estimation module \cite{DBLP:journals/tcsv/ZhangWZC23,zhangVCIP} to aid in motion estimation. Such approaches often result in unnecessary duplication of motion estimation and compensation processes, diminishing overall efficiency. When addressing the second challenge, our experiments reveal that current methods still exhibit significant blurring during large motion events.

\begin{figure}[t]
	\centering
	\includegraphics[width=8cm]{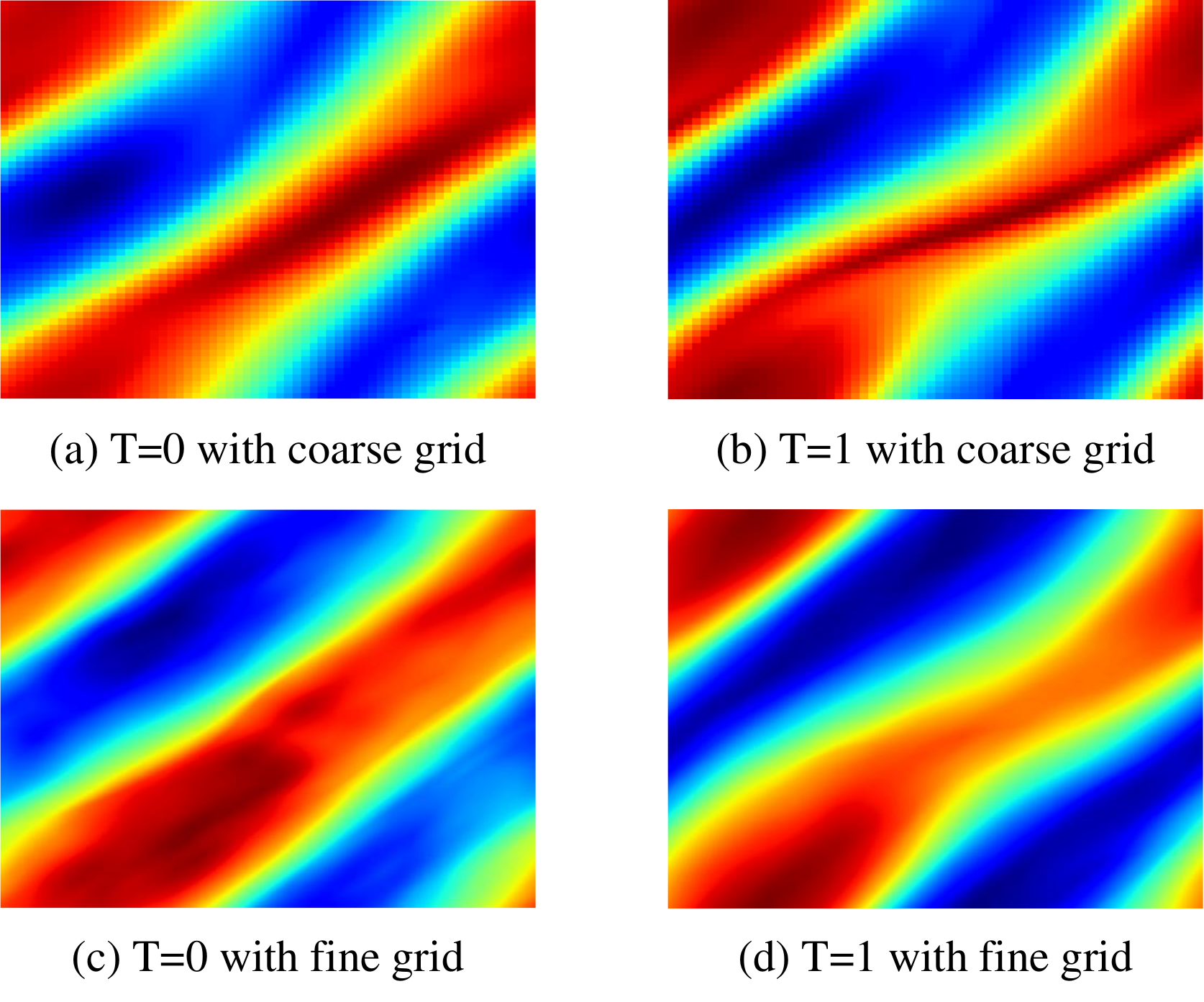}
	\caption{Visualization of the Navier-Stokes equation problem, Physics-Informed Neural Operator (PINN) focuses on two main issues: (1) Sequence prediction problem: Given a time series of fluid fields as input, the neural operator aims to predict fluid changes for the next time interval. (e.g., (a) $\rightarrow$ (b)). (2) Zero-shot super-resolution, which involves training on lower resolution data with coarse-grained discretization and evaluating on higher resolution data with fine-grained discretization. (e.g., Training: (a) $\rightarrow$ (b), evaluation: (c) $\rightarrow$ (d)).) 
	} \label{fig0}
\end{figure}

Neural operator (NO) \cite{DBLP:journals/corr/abs-2003-03485} is a recent innovation in neural networks designed to solve partial differential equation (PDE) problems. The objective of NO is to learn mappings between two infinite-dimensional function spaces, which have been applied broadly in several domains \cite{DBLP:journals/corr/abs-2111-13587,wen2022u,fonseca2023continuous}. Some recent studies have shown that some neural operators can learn the resolution-invariant solution \cite{DBLP:conf/iclr/LiKALBSA21} in the turbulent regime. These observations show remarkable similarities between PDE-solving problems and space-time video super-resolution tasks (As illustrated in Fig.~\ref{fig0}). For instance, the viscosity map in 2D Navier-Stokes equations \cite{constantin1988navier} for predicting turbulent flow can be likened to video frame interpolation. Certain NOs \cite{DBLP:conf/iclr/LiKALBSA21,DBLP:conf/iclr/TranMXO23,DBLP:journals/corr/abs-2301-10022} have been found to possess resolution-independent characteristics, similar to a zero-shot super-resolution.
This insight leads us to reframe the ST-VSR problem. Due to the inherent nature of the STVSR task, it needs to mine fine-grained representations containing rich inter-frame spatio-temporal information from coarse-grained features that only contain intra-frame information. These fine-grained representations will serve as input for the subsequent upsampling stage. Therefore, from the perspective of neural operators, our task is transformed into learning the mapping between two continuous function spaces with different spatio-temporal representation granularities.
Under this framework, we introduce an ST-VSR neural operator (STNO) incorporating a Galerkin-type kernel integral to tackle the challenges above. Benefiting from the linear complexity of Galerkin-type attention, we do not perform any typical patch partition \cite{swinir21,cao2023ciaosr} operators, which are widely adopted in transformer-based methods but directly estimate motion with a global receptive field. This significantly enhances the precision and efficiency of motion estimation, particularly with extreme motion. Moreover, the neural operator's robust modeling capabilities allow for the consolidation of motion information for alignment and interpolation, thereby eliminating redundant MEMC calculations. Our contributions are outlined as follows:
\begin{itemize}
\item We model space-time video super-resolution as a neural operator learning task in two continuous function spaces. The space-time neutral operator (STNO) learner aims at translating coarse-grained spatial data, containing only information within individual frames, into high-quality results that incorporate information across frames.  The improved fine-grained representation can assist in achieving better spatiotemporal restoration.
\item We propose a Galerkin-type attention mechanism for motion estimation and motion compensation (MEMC). Thanks to its linear complexity, we can efficiently perform MEMC with the global receptive field thus improving the accuracy and reliability of motion information, especially in cases involving fast and extreme motions.
\item  Extensive experiments demonstrate the proposed approach outperforms existing methods in both fixed and continuous ST-VSR tasks with faster speed and reduced parameters.
\end{itemize}	
The rest of the paper is organized as follows: Section~\ref{sec:related} reviews the relevant literature. Details of the proposed methodology is
given in Section~\ref{sec:method}. Experiments and analyses are provided
in Section~\ref{exp}. Finally, conclude the paper in Section~\ref{conclu}.
\section{Related Work}
\label{sec:related}
\subsection{Space-Time Video Super-Resolution}
With the rapid development of deep learning, numerous video frame interpolation (VFI)\cite{DAIN,huang2020rife,kong2022ifrnet,DBLP:journals/tip/ShenBZCMG21} and video super-resolution (VSR)\cite{DBLP:journals/tip/LiLW19,2019Multi,2020BasicVSR,DBLP:journals/tip/WenRSNZC22,DBLP:journals/tcsv/LiuJYLZ23} methods have been proposed, continuously pushing the benchmark performance. Readers can refer to the surveys  \cite{dong2023video} and \cite{liu2022video} for more detailed information.
Since both VFI and VSR require the effective utilization of spatiotemporal information from multiple frames, the natural idea is to consider them as a joint task, referred to as space-time video super-resolution. As a representative approach, Zooming Slow-Mo \cite{2021Zooming} establishes a comprehensive framework that utilizes ConvLSTM to aggregate spatiotemporal information. Building upon the Zooming Slow-Mo, Xu\emph{ et al.} \cite{xu2021temporal} introduce a temporal modulation network for controllable intermediate feature interpolation. To further enhance flexibility, recent work has begun to explore continuous space-time video super-resolution. 
Chen \emph{ et al.} \cite{chen2022videoinr} propose an arbitrary spatiotemporal video super-resolution framework called VideoINR, which enables continuous space-time upscaling.  Most recently, MoTIF \cite{chen2023motif} replaces the backward warping module in VideoINR with forward warping, resulting in improved performance.  Although previous approaches have made significant progress, there are still issues with the efficiency of multi-frame motion estimation and motion compensation, particularly in handling large motions, which can lead to problems such as blurry artifacts and temporal inconsistency.
\subsection{Neural Operators}
Neural operator \cite{DBLP:journals/corr/abs-2003-03485} (NO) is first introduced to solve partial differential equation (PDE) problems. It approximates the mapping between two infinite-dimensional spaces with the composition of nonlinear activation functions and a specific class of integral operators.
Based on previous work \cite{DBLP:journals/corr/abs-2003-03485}, the Fourier neural operator (FNO) is proposed by parameterizing
the integral kernel directly in Fourier space. By utilizing the Fourier transform, FNO can model long-range information and capture spatial information over long distances with quasi-linear complexity.
Cao \emph{et al.} \cite{DBLP:conf/nips/Cao21} introduce a novel integral operator that incorporates a modified attention mechanism. This Galerkin-type attention explicitly represents a Petrov-Galerkin projection and exhibits linear complexity attention without the softmax operation. Most recently, Wei \emph{et al.} \cite{wei2023super} propose utilizing NO to address the task of single image super-resolution. Although this method has demonstrated its superiority, exploring more efficient utilization for multi-frame image restoration with neural operators still needs to be explored.
In reality, NOs are primarily employed to tackle time-related prediction problems; utilizing them solely to enhance intra-frame information fails to exploit the potential of neural operators fully. In light of this, our paper focuses on exploring more efficient and effective inter-frame MEMC.
\section{Methodology}
\label{sec:method}
\subsection{ Problem Formulation}
\label{prob1}
\begin{figure*}[!t]
	\centering
	\includegraphics[width=17cm]{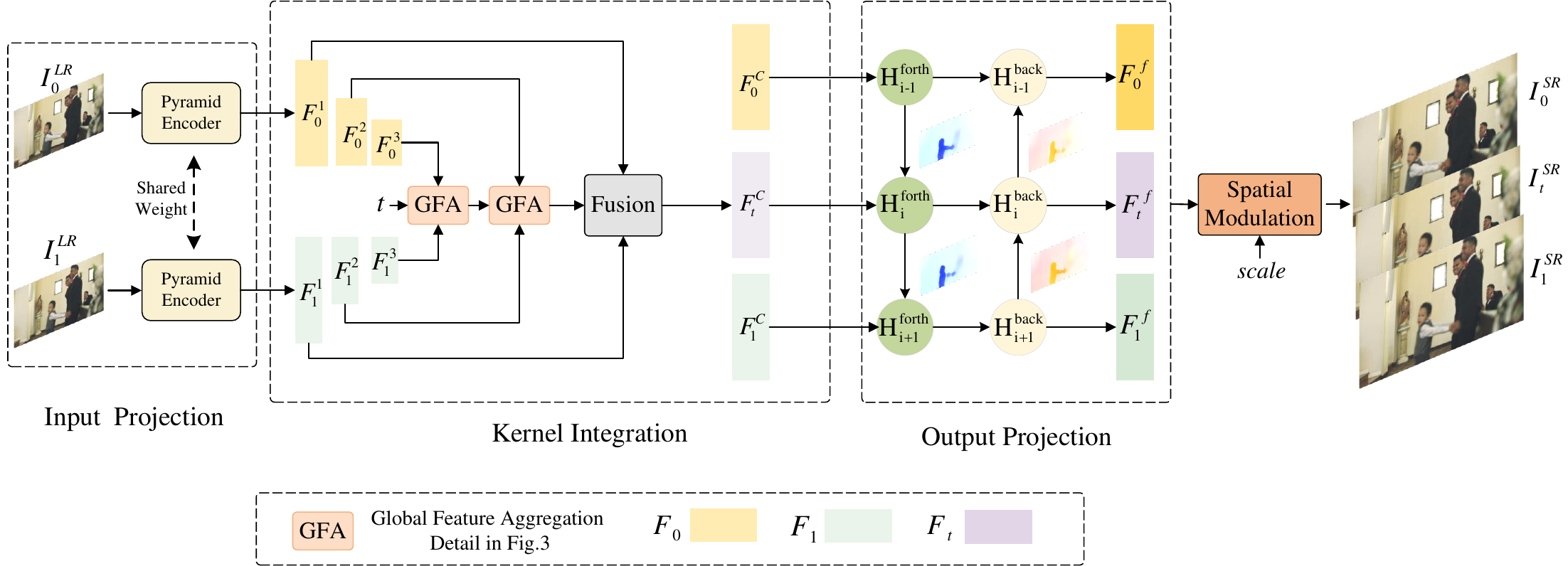}
	\caption{Overview of the proposed method. We first extract multi-scale coarse-grained intra representations  $F_{\{0,1\}}^{\{1,2,3\}}$, which a kernel-integrated operator subsequently processes to perform multi-scale MEMC. The obtained coarse-grained features $F^{C}$ and motion information are further enhanced through multi-frame information propagation, resulting in fine-grained representations $F^{f}$, which contain rich spatiotemporal information. Finally, the obtained $F^{f}$ are used for upsampling.
	} \label{fig1}
\end{figure*}
The concept of the neural operator was initially introduced in \cite{DBLP:journals/corr/abs-2003-03485} intending to learn a mapping between infinite-dimensional spaces using a finite set of input-output data pairs.
Generally, a typical neural operator consists of three components, and we borrow the language from \cite{DBLP:conf/iclr/LiKALBSA21} to provide a brief introduction. 1) Input projection: This component typically serves as a task-specific feature extractor that maps the data into a higher-dimensional space. 2) Kernel integral operator: This component extracts information from the input data, often exhibiting temporal correlations. As a result, sequential data modeling problems (e.g., Burgers' equation, Navier-Stokes equation) are commonly associated with transformers \cite{atten17}. \\
3) Output projection: This component maps the data to the output feature space. 
In our work, the proposed STNO also consists of these three components, and each component has been designed with task-driven considerations.
In contrast to PDE solving tasks that have well-defined physical backgrounds \cite{DBLP:conf/iclr/LiKALBSA21,ren2022phycrnet}, directly learning the operator from low resolution and low frame rate (LRLF) video to high resolution and high frame rate (HRHF) video is exceptionally challenging due to the highly nonlinear nature of real-world motion. We need to explore rich inter-frame spatiotemporal information from a video sequence for video super-resolution tasks. Therefore, we are not directly learning the mapping from LRLF to HRHF but rather from the input's coarse representations 
$F^{c}$ to accurate representations $F^{f}$ containing abundant spatiotemporal information. 
Let  $D \subset \mathbb{R}^d$ be a bounded, open set and $ \mathcal{A} = \mathcal{A}(D;\mathbb{R}^{d_a})$ and $ \mathcal{U} = \mathcal{U}(D;\mathbb{R}^{d_u})$ be separable Hilbert spaces $\mathcal{H}$ of function taking values in $\mathbb{R}^{d_a}$ and $\mathbb{R}^{d_u}$.
The input LRLF features and the corresponding fine-grained outputs are vector-valued functions within  $\mathcal{A}$  and  $\mathcal{U}$, respectively.
To work with them numerically, we access the discretized function values of input video at $(x, y, t)$ through their corresponding spatial and temporal coordinate. Given sampled function pairs $\{a^{k},u^{k}\}_{k=1}^{N}$ where $a^{k}$  $\sim$ $\mu$ is an i.i.d. sequence from the probability measure $\mu$ supported on  $\mathcal{A}$, we introduce a neural operator $\mathcal{G}^{\dagger}: \mathcal{A} \rightarrow \mathcal{U} $, parametrized by $\theta$, that aims to learn the mapping between two infinite dimensional space. 
In practice, the process of operator learning $\mathcal{G}^{\dagger} \leftarrow \mathcal{G}_{\theta} $ is often modeled as an empirical risk minimization problem which is the minimizer of discretized sampled observations $u^{k}$ and $a^{k}$ with the cost function $C$:
\begin{equation}
	\begin{aligned}
		&\min_{\theta}{\mathbb{E}_{a\sim\mu}\lVert C(\mathcal{G}(a,\theta)-\mathcal{G}^{\dagger}(a))\rVert}, \\
		&\approx \min_{\theta}\frac{1}{N}\sum_{k=1}^N{\lVert u^{(k)} - \mathcal{G}_{\theta}(a^{(k)})\rVert}_{\mathcal{U}},
	\end{aligned}
\end{equation}
where $a$ and $u$ denote LRLF input features and corresponding representation with abundant spatiotemporal information. 
As \cite{DBLP:journals/corr/abs-2003-03485,DBLP:conf/iclr/LiKALBSA21} shows, the operator $\mathcal{G}$ is often modeled as an iterative architecture
with dimension $d_z$. Then $\mathcal{G}_{\theta}: \mathcal{A} \rightarrow \mathcal{U}$ can be formulated as follows:
\begin{equation}
	\begin{aligned}
		%\mathcal{G}_{i} v_{i} (x) = \sigma(\int k_{i}(x,y)v_{i} d {\mu}_{i}(y)+W_{i}v_{i}(x) )
		v_{0}(x) &= \mathcal{P}(x,a(x)),\\
		v_{t+1}(x) &= \sigma(W_{t}v_{t}(x)+\mathcal{K}_{t}(v_{t};\theta)(x)) \label{eq5}, \\
		u(x) &= \mathcal{Q}(v_{T}(x)),
	\end{aligned}
\end{equation}
\begin{figure*}[!t]
	\centering
	\includegraphics[width=16cm]{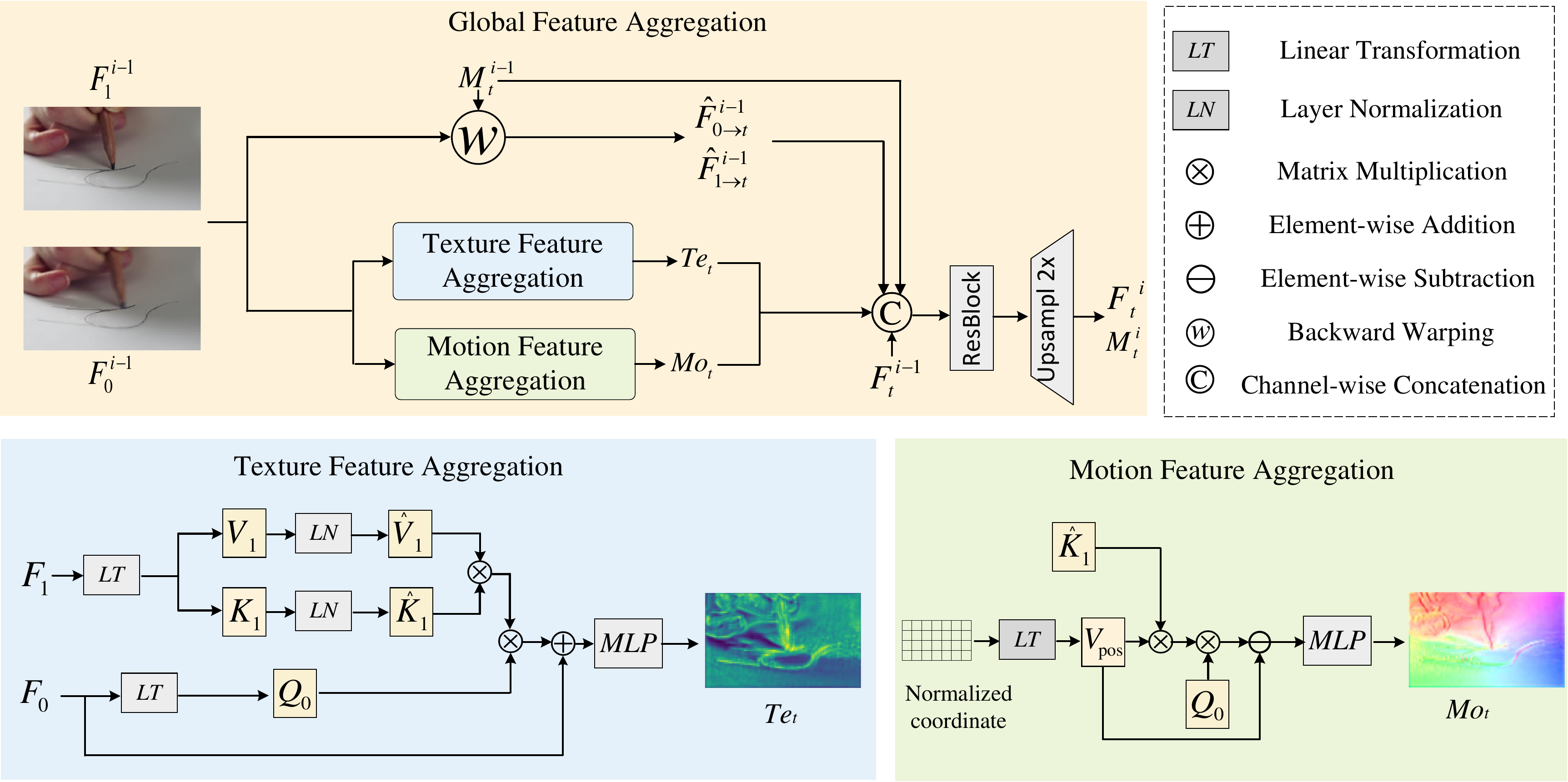}
	\caption{
		The Global Feature Aggregation module comprises two main components: Texture Feature Aggregation and Motion Feature Aggregation. We employ a Galerkin-type attention mechanism to capture global texture features $Te_{t}$ and motion features $Mo_{t}$. The generated $Te_{t}$ and  $Mo_{t}$ are coupled together, mutually enhancing each other, ultimately resulting in high-quality interpolated intermediate frame features and motion flow.
	} \label{fig2}
\end{figure*}
where $\mathcal{P}: \mathbb{R}^{d_a+d} \rightarrow \mathbb{R}^{d_z}$, and $\mathcal{Q}: \mathbb{R}^{d_z} \rightarrow \mathbb{R}^{d_u}$ are the input and output projection functions respectively, mapping the input $a$ to its hidden representation $v_0$ and the last layer hidden representation $v_T$ back to the output function $u$. $W: \mathbb{R}^{d_z} \rightarrow \mathbb{R}^{d_z}$ is a point-wise linear transformation, and $ \sigma: \mathbb{R}^{d_z} \rightarrow \mathbb{R}^{d_z}$ is the nonlinear activation function.
$\mathcal{K}(a;\theta)$ is a kernel integral transformation parameterized by a neural network, mapping to the bounded operators on $\mathcal{U}(D,\mathbb{R}^{d_{z}})$.
And the kernel integral operator in Eq.~\ref{eq5} can be formulated as:
\begin{equation}
	\begin{aligned}
		(\mathcal{K} (a;\phi)v_{t})(x) &= \int_{D} \mathcal{K} (x,y,a(x),a(y);\phi)v_t(y) \dd y ,\\ \forall &x \in D ,
	\end{aligned}\label{eq2}
\end{equation}
where $\mathcal{K}_{\phi}: \mathbb{R}^{2(d+{d_a})} \rightarrow \mathbb{R}^{d_{z}\times d_{z}} $ is a neural network with learnable parameter $\phi \in {\Theta}_{\mathcal{K}}$, $a(x),a(y)$ stands for the discrete sampled points corresponding to the spatial positions $(x, y)$.
With the learned transformation $\mathcal{G}$, the coarse-grained representation corresponding to LRLF can be converted into fine-grained representations containing rich spatiotemporal information. And the next
step is to modulate the fine-grained features using a learnable
underlying function to the desired HRHF outputs.

\subsection{Network Architecture}
After the problem formulation, we introduce the specific network architecture, which is divided into three stages corresponding to Section~\ref{prob1}.
Consider a video sequence consisting of $N$ low-resolution images, which can be represented as a discretized vector-valued function $\mathcal{H} \ni f : \Omega \rightarrow \mathbb{R}^{N \times H \times W\times 3 }$ containing only independent intra-frame information. We aim to use neural operators to extract fine-grained spatiotemporal representations that incorporate abundant inter-frame information.
The framework of the proposed method is provided in Fig.~\ref{fig1}. \\
\noindent \textbf{Input Projection.}
Following some recent NO methods \cite{DBLP:conf/nips/Cao21,ren2022phycrnet}, we first employ a stack of residual layers as a feature extractor to map the input data to a higher-dimensional space. Additionally, to capture multi-scale motion, we employ convolutions with a stride of 2 to obtain feature maps at three different resolutions concerning $I_{0}$ and $I_{1}$, which we denote as $F_{\{i\}}^{\{j\}}$ ($ i \in \{0,1\}$,$j \in \{ 1,2,3\}$).  \\
\noindent \textbf{Kernel Integration.}
After obtaining the multi-scale features, we proceed to the kernel integration stage, where the goal is to extract motion information between adjacent frames and interpolate intermediate frames. 
Specifically, given the features $F_{0}$ and $F_{1}$ corresponding to the adjacent two frames, as well as the intermediate time step $t$, we aim to obtain the intermediate frame features $F_{t}$ and the corresponding inter-frame motion information $M_{t}$. This process can be represented as:
\begin{equation}
	\begin{aligned}
		M_{t},F_{t} = \mathcal{K}(F_{0},F_{1},t),
	\end{aligned}
\end{equation}
where $\mathcal{K}$

represents the underlying kernel integral operator in Eq.~\ref{eq2}, and $M_{t}$ represents the inter-frame motion information. Specifically, since our task is spatiotemporal super-resolution, we not only need the intermediate motion information $ flow_{t \rightarrow 0,1}$ to fit the temporally interpolated frames but also require the motion information between pre-existing frames $ flow_{0 \rightarrow 1,1 \rightarrow 0}$ for feature alignment.

Motion estimation for low-level vision tasks has been extensively studied for years. One persistent challenge is estimating large motions. The current state-of-the-art solutions often employ transformer \cite{lu2022video,zhang2023extracting} architectures. However, vision transformers commonly suffer from quadratic computation and memory costs.
Although patch partition operations \cite{DBLP:conf/iclr/DosovitskiyB0WZ21,swinir21} can reduce complexity to some extent, they inevitably lead to limited receptive fields and suffer from blocking artifacts \cite{DBLP:conf/cvpr/ChenWZ0D23}.
Inspired by recent advances in NO, we propose a global feature estimation module with linear complexity to address this issue, and the illustration is given in Fig.~\ref{fig2}.
Following the design of transformers, we use three learnable linear transformation matrices $W_{q}$, $W_{k}$, and $W_{v}$ to multiply with the features to obtain the corresponding Query, Key, and Value:

\begin{equation}
	\begin{aligned}
		Q_{0} = F_{0} \cdot W_{Q},
		K_{1} = F_{1} \cdot W_{K},
		V_{1} = F_{1} \cdot W_{V}.
	\end{aligned}
\end{equation}
Our goal is to use $Q_{0}$ to query $K_{1}$ and obtain similar appearances and motion information between $I_{0}$ and $I_{1}$.
To efficiently capture large motion, inspired by \cite{DBLP:conf/nips/Cao21} we introduce a Galerkin-type attention mechanism that does not require softmax. It effectively replaces the traditional attention mechanism ($ atten = softmax(\frac{Q K^{T}}{\sqrt{d_{z}}})V $) and has linear computational complexity.

Consider an operator learning problem with an underlying domain $\Omega \subset \mathbb{R}^{n \times d_{z}}$ , $\{{x}_{i}\}_{i=1}^{n} $ denotes the set of feature points in the discretized $\Omega$.
The columns of $Q$/$K$/$V$, respectively, contain the vector representations of the learned basis functions spanning certain subspaces of the latent representation Hilbert spaces. Following settings in \cite{DBLP:conf/nips/Cao21}, we assume $Q$,$K$, and $V$ are defined on the same spacial domain $\Omega$ with $d_{z}$ dimensional vector-valued functions, with its subscript denoting the corresponding component. $k_{l}(\cdot)$, $v_{j}(\cdot)$ denote the vector representations of $k_{l} (1 \le l \le d_z)$ and $v_{j}$ that evaluated at every sampled $x_{i}$. The kernel integral attention can be formulated as:
\begin{equation}
	\begin{aligned}
		(\mathcal{K}(z)(x))_{j} = \sum_{l=1}^{d_z} \langle v_{j}, k_{l}\rangle q_{l}(x) \quad for \quad j = 1,\dots ,d, \\ \approx \sum_{l=1}^{d_z} \left( \int_{\Omega}  v_j(\xi) k_l(\xi) \dd \xi \right) q_l(x_i), \forall x \in \Omega,
	\end{aligned}
\end{equation}
in other words, this Galerkin-type attention of the output $z$ can be then compactly written as:
\begin{equation}
	\label{eq7}
	\begin{aligned}
		z = \frac{Q_{0}(\hat{K}_{1}^{T}\hat{V}_{1})}{n}.
	\end{aligned}
\end{equation}
In terms of the specific implementation process, 
given  $Q_{0},K_{1},V_{1} \in \mathbb{R}^{H \times W \times C}$, the texture feature $Te_{t}$ is calculated as:
\begin{equation}
	\label{te}
	\begin{aligned}
		Te_{t} = MLP(F_{0} + Q_{0}  (\hat{K}^{T}_{1} \hat{V}_{1})/(H \times W)),
	\end{aligned}
\end{equation}
where $\hat{K}_{1} = LN(K_{1})$,$\hat{V}_{1} = LN(V_{1})$, and $LN(\cdot)$ denotes layer normalization operator, and MLP denotes the Multilayer Perceptron.
Unlike traditional attention mechanisms with quadratic complexity ($O({(HW)}^{2}C)$), this Galerkin-type attention exhibits only linear complexity ($O((HW) C^{2}$)), making it highly efficient. The detailed proof process can be found in \cite{DBLP:conf/nips/Cao21}.
Considering the alignment capability of the transformer \cite{DBLP:conf/nips/ShiGXWYD22} itself, $Te_{t}$ actually integrates similar regions from $I_{0}$ and $I_{1}$ and adaptively fits the intermediate feature corresponding to time $t$.
Like the process of modeling texture information, we model the motion information $M_{t}$ similarly. First, we feed the normalized coordinate information into an MLP to modulate the positional information. Then, we estimate the motion information from $I_{0}$ to $I_{1}$ using inter-frame attention. The  motion feature $Mo_{t}$ is obtained by subtracting the original positional information from the motion representation that aggregates the pixel positional relationships between the two frames:
\begin{equation}
	\begin{aligned}
		Mo_{t} &=  MLP(Q_{0}(\hat{K}^{T}_{1} V_{pos})/(H \times W) - V_{pos}),  
	\end{aligned}\label{eq9}
\end{equation}
where $V_{pos}$ represents the normalized positional coordinates.
Simultaneously, to gather more information and allow mutual enhancement between motion and texture, we perform both explicit (feature warping) and implicit (kernel integer operator) MEMC. As shown in Fig.~\ref{fig2}, the feature aggregation and motion estimation are conducted in a coarse-to-fine pyramid manner. It means that the downsampled low-resolution features are first used to estimate coarse motion at smaller scales and gradually upsampled. The estimation results at the ${(i-1)}^{th}$ layer serve as guidance for the $i^{th}$ layer, aligning with the iterative architecture of NO and the principle of residual learning. \\
\noindent \textbf{Output Projection.}
After the kernel integral operation, we obtain temporally interpolated feature $F_{t}$, as well as the inter-frame motion flow $M_{t}$ ($flow_{0 \rightarrow 1}$, $flow_{1 \rightarrow 0}$, $flow_{t \rightarrow 0}$, and $flow_{t \rightarrow 1}$). 
The next step involves alignment within multiple frames to ensure that each frame receives supplementary information from the others (commonly called temporal propagation in video super-resolution). Here, we employ the popular bidirectional recurrent structure \cite{2020BasicVSR} to facilitate global propagation. Given a feature $x_{i}$ corresponding to the $i^{th}$ input LR, and the  features
propagated from its neighbors, denoted as $H^{f}_{i-1}$ and $H^{b}_{i+1}$, we have:
\begin{equation}
	\begin{aligned}
		H_{i}^{b} &= P_{b}(x_{i},x_{i+1},H^{b}_{i+1}), \\
		H_{i}^{f} &= P_{f}(x_{i},x_{i-1},H^{f}_{i-1}), \\
		\text{where}  & \quad H_{i}^{b,f} = \mathcal{W}(H^{b,f}_{i\pm 1},M^{b,f}_{i}),
	\end{aligned}
\end{equation}
where $P_{b}$ and $P_{f}$ denote the backward and forward propagation branches, and $\mathcal{W}$ denotes backward warping. 
For the temporally interpolated frames, since the corresponding motion information $flow_{t \rightarrow 0}$ and $flow_{t \rightarrow 1}$ have also been computed, feature propagation is carried out similarly as with the originally available frames. Now, the feature representations for all frames contain rich spatiotemporal information and are ready for upsampling. \\
\noindent \textbf{Spatial Modulation.} The final step is to decode the feature as  RGB values.
To modulate the fine-grained spatiotemporal representations to arbitrary scales, we adopt a method similar to \cite{DBLP:conf/cvpr/ChenL021}, where an MLP is used to predict the RGB values for each spatial position. Specifically, given the fine-grained spatiotemporal representation $R_{i}$ corresponding to $i^{th}$ frame and the associated coordinate position information $coord$, the super-resolution process can be represented as follows:
\begin{equation}
	\begin{aligned}
		SR_{i} = MLP_{s}(R_{i},coord),
	\end{aligned}
\end{equation}
where $MLP_{s}$ denotes the underlying spatial modulation function.
It is worth noting that, due to the powerful modeling capability of the neural operator, we only adopt local ensembling to aggregate the four nearest feature positions without the need for additional local feature unfolding operations \cite{DBLP:conf/cvpr/ChenL021} or feature aggregation \cite{chen2022videoinr,chen2023motif} process. It significantly reduces computational complexity and improves speed.
\section{Experiments}
\label{exp}
\subsection{  Experiments Setup}
\begin{table*}[!t]
	\caption{Quantitative comparisons of PSNR (dB), SSIM, speed (FPS), on Vid4, Vimeo-90K-T, GoPro and Adobe. The
		inference time is calculated on Vid4 dataset with one Nvidia 1080Ti GPU. 
		$\dagger$ denotes only utilizing two adjacent images. 
		For fairness, we calculate the speed using STNO-fix-two that takes two frames as input at a time.
		The best two results are highlighted
		in red and blue colors.}
	\resizebox{1.01\linewidth}{!}{	
		\begin{tabular}{c|cc|cc|cc|cc|cc|cc|c|c}
			\hline
			STVSR / VFI+VSR       & \multicolumn{2}{c|}{\begin{tabular}[c]{@{}c@{}}Vid4\\ PSNR SSIM\end{tabular}} & \multicolumn{2}{c|}{\begin{tabular}[c]{@{}c@{}}Vimeo-Fast\\ PSNR SSIM\end{tabular}} & \multicolumn{2}{c|}{\begin{tabular}[c]{@{}c@{}}Vimeo-Medium\\ PSNR SSIM\end{tabular}} & \multicolumn{2}{c|}{\begin{tabular}[c]{@{}c@{}}Vimeo-Slow\\ PSNR SSIM\end{tabular}} & \multicolumn{2}{c|}{\begin{tabular}[c]{@{}c@{}}GoPro\\ PSNR SSIM\end{tabular}} & \multicolumn{2}{c|}{\begin{tabular}[c]{@{}c@{}}Adobe\\ PSNR SSIM\end{tabular}} & \multicolumn{1}{l|}{\begin{tabular}[c]{@{}c@{}}Speed\\ FPS\end{tabular}} & \multicolumn{1}{l}{Parameter} \\ \hline
			SuperSloMo\cite{0Super}+Bicubic  & 22.84                                 & 0.5772                                & 31.88                                    & 0.8793                                   & 29.94                                     & 0.8477                                    & 28.73                                    & 0.8102                                   &         27.50                              &         0.8094                 &     26.51                                  &   0.7650                                     &                          - &19.8                               \\
			SuperSloMo\cite{0Super}+EDVR\cite{2019EDVR}    & 24.40                                 & 0.6706                                & 35.05                                    & 0.9136                                   & 33.85                                     & 0.8967                                    & 30.99                                    & 0.8673                                   &     29.75                                  &          0.8589              &       28.97                                &               0.8398                         &                         4.94 & 19.8+20.7                              \\ \hline
			DAIN\cite{DAIN}+Bicubic        & 23.55                                 & 0.6268                                & 32.41                                    & 0.8910                                   & 30.67                                     & 0.8636                                    & 29.06                                    & 0.8289                          &  28.41  & 0.8299     &   27.35      &   0.7877               &                          -&24.0                               \\
			DAIN\cite{DAIN}+EDVR\cite{2019EDVR}           & 26.12                                 & 0.7836                                & 35.81                                    & 0.9323                                   & 34.66                                     & 0.9281                                    & 33.11                                    & 0.9119                                   &   31.21                                    &              0.8904                          &       30.80                                &    0.8871       &                         4.00 &24.0+20.7                               \\
			DAIN\cite{DAIN}+BasicVSR\cite{2020BasicVSR}       &  25.79                          &     0.7711                                  &              35.07                            &      0.9275                                    &       33.82                                    &      0.9211                                     &         32.08  &     0.8988          &      31.01   &  0.8878    &  30.59    &              0.8836           &                          16.72&24.0+6.3                               \\ \hline
			EDSC\cite{EDSC}+BasicVSR++\cite{chan2021basicvsr++}     & 26.27                                 & 0.7900                                & 35.80                                    & 0.9315                                   & 34.62                                     & 0.9277                                    & 33.20                                    & 0.9129                                   & 31.09                                      &  0.8880                                      & 30.68                                      & 0.8823                                       &                          10.31&8.9+7.3                               \\
			IFRnet\cite{kong2022ifrnet}+BasicVSR++\cite{chan2021basicvsr++}   & 26.36                                 & 0.7917                                & 36.26                                    & 0.9358                                   & 34.80                                     & 0.9292                                    & 33.22                                    & 0.9126                                   & \textcolor{blue}{31.31}                                   &  0.8913                                      & \textcolor{blue}{30.89}                                      & 0.8861                                       &                          8.93&19.7+7.3                               \\
			RIFE\cite{huang2020rife}+BasicVSR++\cite{chan2021basicvsr++}     & 26.37                                 & 0.7939                                & 36.01                                    & 0.9337                                   & 34.64                                     & 0.9275                                    & 33.19                                    & 0.9122                                   & 31.17                                      & 0.8891                                       &  30.75                                     &  0.8833                                      &                         7.25 &20.9+7.3                               \\ \hline
			Zooming SlowMo\cite{2021Zooming}      & 26.31                                 & 0.7976                                & 36.81                                    & 0.9415                                   & 35.41                                     & 0.9361                                    & 33.36                                    & 0.9138                                   & 30.57                                      &   \textcolor{blue}{0.9031}                                     & 30.29                                      & \textcolor{blue}{0.8996}                                       &                          \textcolor{blue}{12.40}&11.1                               \\
			TMnet\cite{xu2021temporal}              & \textcolor{blue}{26.43}                                 & \textcolor{blue}{0.8016}                                & \textcolor{blue}{37.04}                                    & \textcolor{blue}{0.9435}                                   & \textcolor{blue}{35.60}                                     & \textcolor{blue}{0.9380}                                   & \textcolor{blue}{33.51}                                   & \textcolor{blue}{0.9159}                                   &   30.37                                    &0.9027                                        & 30.12                                      &   0.8984                                     &                         11.60 & 12.3                              \\
			VideoINR$\dagger$\cite{chen2022videoinr}            & 25.78                                 & 0.7753                                & 33.95                                    & 0.9232                                   & 33.37                                     & 0.9214                                    & 31.71                                    & 0.8963                                   &30.73                                       &0.8850                                        & 30.21                                      &0.8805                                        &                         0.77 & 11.3                              \\
			MoTIF$\dagger$\cite{chen2023motif}               & 25.79                                      & 0.7745                                      & 35.01                                         & 0.9281                                         & 33.98                                          &0.9235                                           & 32.11                                         & 0.8981                                         & 31.04                                      &   0.8877                                     &  30.63                                     & 0.8839                                      &                         0.52 & 12.6                              \\ \hline
			STNO-fix-two$\dagger$ (Ours)               & 25.94                                      & 0.7773                                      & 36.30                                         & 0.9365                                         & 34.80                                          &0.9291                                           & 32.87                                         & 0.9060                                         & 31.20                                      &   0.8901                                     &  30.83                                     & 0.8865                                      &                          \textcolor{red}{17.23}&9.37                               \\ 
			STNO-fix (Ours)                & \textcolor{red}{26.73}                                 & \textcolor{red}{0.8138}                               & \textcolor{red}{37.13}                                    & \textcolor{red}{0.9447}                                 & \textcolor{red}{35.61}                                    & \textcolor{red}{0.9383}                                    & \textcolor{red}{33.52}                                   & \textcolor{red}{0.9160}                                   & \textcolor{red}{32.06}                                & \textcolor{red}{0.9060}                                & \textcolor{red}{31.85}                                 & \textcolor{red}{0.9073}                                & -                    & \textcolor{red}{9.37}                          \\ \hline
	\end{tabular}		}\label{tab1}
\end{table*}
\begin{figure*}[h]
	\centering
	\includegraphics[width=15cm]{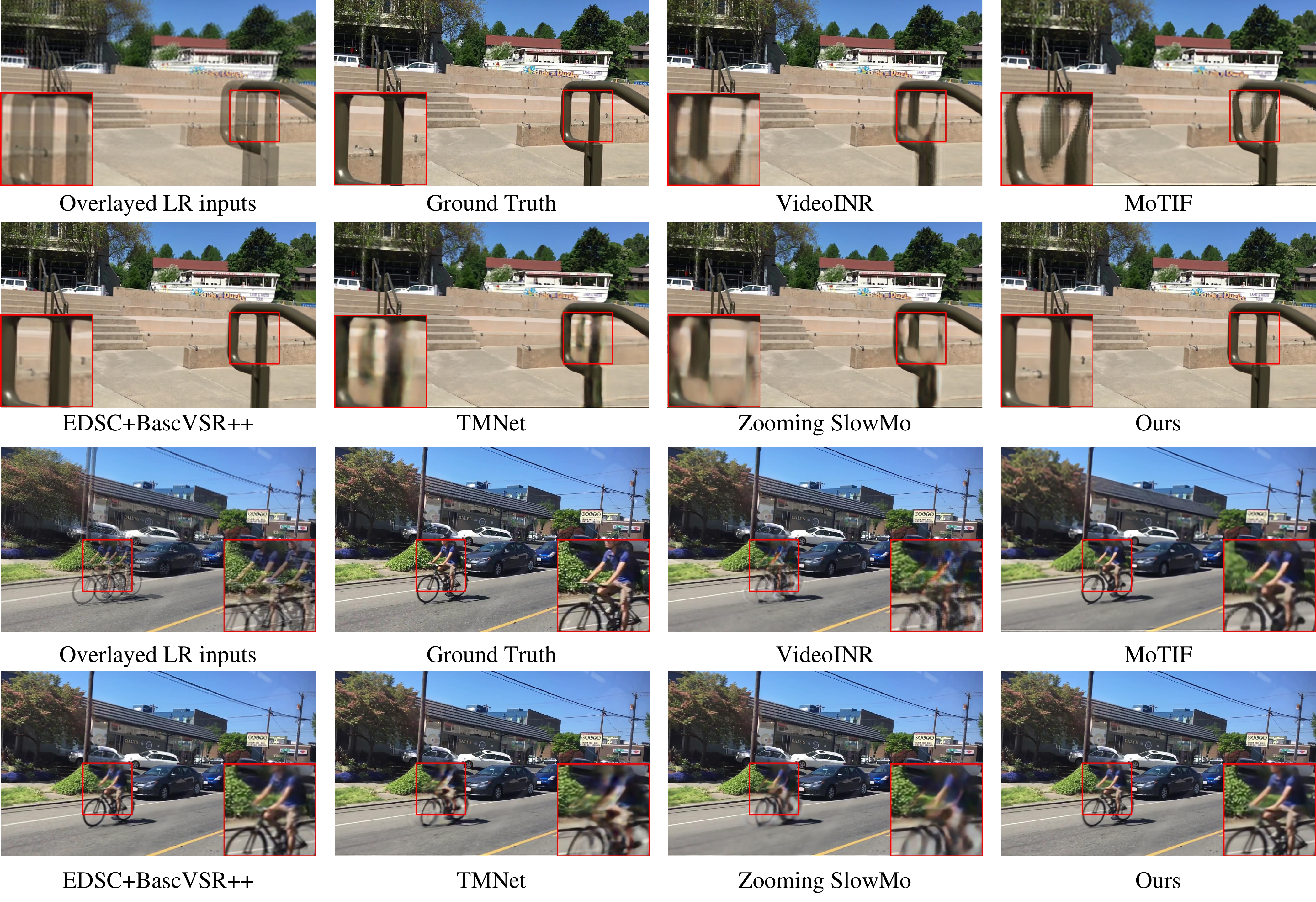}
	\caption{Synthetic intermediate frame by different methods for large motion on Adobe. Pay attention to the areas outlined in red boxes, and zoom in for a better view. } 
	\label{fig4}
\end{figure*}
\noindent \textbf{Datasets.} 
We use multiple datasets to train and test our model. The datasets we used are as follows:\\
\noindent Vime90k-T\cite{DBLP:journals/corr/abs-1711-09078}: This dataset comprises 91,701 video clips, with each clip consisting of seven consecutive frames at a resolution of 448 $\times$ 256. Following the approach in \cite{2021Zooming,xu2021temporal}, we divided the test set of Vimeo-90K-T into three subsets based on the average motion magnitude: fast, medium, and slow motion subsets. \\
\noindent Vid4\cite{5995614}: This dataset consists of four sequences and is characterized by rich textures. It is frequently used as a benchmark for video super-resolution. \\
\noindent Adobe\cite{SuDWSHW17}: It includes 17 test sequences  captured at 240fps with an iPhone
6s. To align with the setup of VideoINR\cite{chen2022videoinr}, we set the size of the temporal sliding window to 8 (i.e., 1$^{st}$, 9$^{th}$, 17$^{th}$, etc.) to generate the input LR frames and use them for interpolating the intermediate frames. \\
\noindent GoPro\cite{DBLP:conf/cvpr/NahKL17}: It contains 11 videos of street scenes captured at high frame rates, presenting challenges associated with both object and camera motion. \\
\noindent SPMCS\cite{tao2017detail}: This dataset includes 32 videos and is widely used as a benchmark for video super-resolution. SPMCS exhibits rich textures and sensitivity to different scaling factors, making it suitable for evaluating the performance of continuous ST-VSR. \\
\noindent \textbf{Implementation Details.} 
We train two models to compare their performance for fixed spatiotemporal upsampling and continuous space-time super-resolution, which we refer to as "STNO-fix" and "STNO-c". 
To train STNO-fix, we crop the images into 256$\times$256 patches as the ground truth and downsample them by a factor of 4 using bicubic interpolation. The resulting frames, specifically the odd frames (e.g., 1$^{st}$, $3^{rd}$, $\dots$), are used as inputs to train the model.
For model optimization, we employ the Adam optimizer \cite{kingma2014adam} with $\beta_1$=0.9 and $\beta_2$=0.999. We also apply standard augmentation techniques, such as rotation, flipping, and random cropping. The initial learning rate was set to 2 $\times$ 10$^{-4}$ 
and decayed to 1 $\times$ 10$^{-7}$ using a cosine annealing scheduler. To facilitate longer information propagation, we adopted the approach described in \cite{2020BasicVSR}, which involves applying temporal augmentation by flipping the original input sequence.

For the STNO-fix, the Vimeo-90K-T dataset \cite{DBLP:journals/corr/abs-1711-09078} is employed as the training set. We apply bicubic downsampling by a factor of 4 on odd-numbered frames (e.g., $1^{st}$,$3^{rd}$ ...) as input, and corresponding high-resolution images (e.g., $1^{st}$,$2^{nd}$,$3^{rd}$ ...) are used as supervision.
We supervise both the pixel-wise predicted RGB pixel values using the Charbonnier loss \cite{lai2017deep} and the predicted motion flow simultaneously to better capture precise motion.
The pseudo-labels for motion flow are obtained using a pre-trained optical flow model \cite{JiangCLLH21}.
\begin{table*}[!t]
	\caption{Quantitative comparison (PSNR (dB)/SSIM) of continuous space-time super-resolution. The results are
		calculated on SPMCS. The best results are highlighted
		in red.}
	\centering
	\resizebox{0.78\linewidth}{!}{	
		\begin{tabular}{cc|cccc}
			\hline
			Time Scale       & Space Scale         & TMnet & VideoINR             & MoTIF         & STNO-c (Ours)           \\         \hline
			$\times$2         & $\times$2          &  -    &30.67/0.9020          &35.64/0.9663  &\textcolor{red}{38.55}/\textcolor{red}{0.9744}     \\
			$\times$2         & $\times$3          &  -     &31.17/0.8959          &31.38/0.9135  &\textcolor{red}{33.38}/\textcolor{red}{0.9245}      \\
			$\times$2         & $\times$4          & 29.23/0.8741     &28.69/0.8430          &28.94/0.8497  &\textcolor{red}{30.58}/\textcolor{red}{0.8620}      \\
			$\times$2         & $\times$5          &  -    &28.05/0.7894          &27.42/0.7934  &\textcolor{red}{28.73}/\textcolor{red}{0.8060}      \\
			$\times$2         & $\times$6          &  -     &26.06/0.7361          &26.33/0.7458  &\textcolor{red}{27.42}/\textcolor{red}{0.7579}     \\
			$\times$4         & $\times$2          &  -    &30.10/0.8904          & 34.59/0.9470 &\textcolor{red}{36.62}/\textcolor{red}{0.9625}      \\
			$\times$4         & $\times$3          &  -    &30.75/0.8870          & 31.17/0.9085 &\textcolor{red}{32.36}/\textcolor{red}{0.9127}      \\
			$\times$4         & $\times$4          &  28.02/0.8265     &28.51/0.8373          & 28.88/0.8482 &\textcolor{red}{29.91}/\textcolor{red}{0.8512}      \\
			$\times$4         & $\times$5          &  -    &27.94/0.7869          & 27.42/0.7942 &\textcolor{red}{28.30}/\textcolor{red}{0.7975}      \\
			$\times$4         & $\times$6          &  -    &26.03/0.7359          & 26.37/0.7484 &\textcolor{red}{27.13}/\textcolor{red}{0.7517}     \\ \hline
		\end{tabular}
	}\label{tab2}
\end{table*}
\begin{figure*}[t]
	\centering
	\includegraphics[width=15cm]{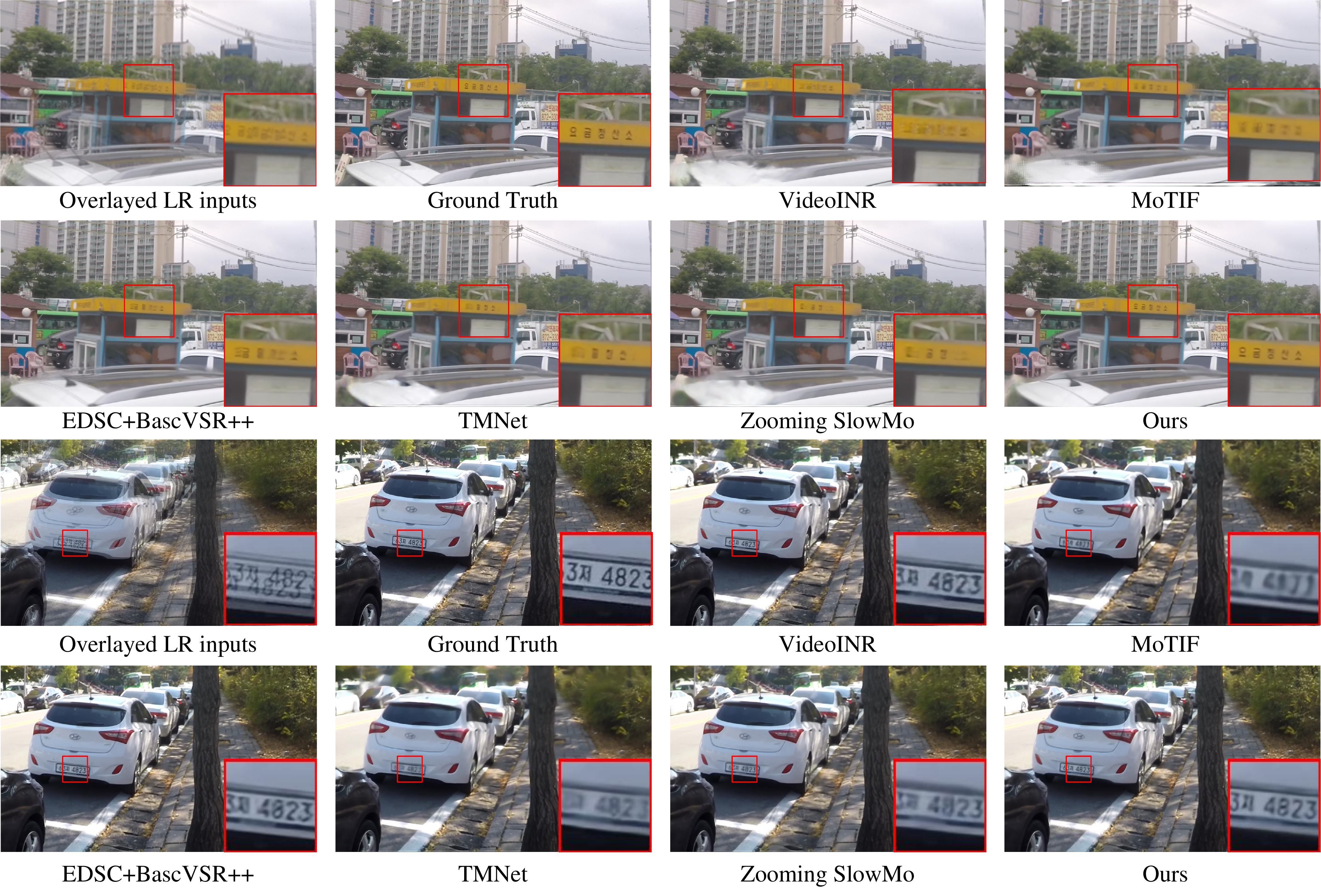}
	\caption{Synthetic intermediate frame by different methods for large motion on Adobe. Pay attention to the areas outlined in red boxes, and zoom in for a better view. } 
	\label{fig5}
\end{figure*}
The loss function can be represented as follows:
\begin{equation}
	\begin{aligned}
		\mathcal{L} = \mathcal{L}_{char}(\hat{I}^{SR}_{t},I_{t}^{HR}) + \alpha \sum_{k=1}^{3} \mathcal{L}_{char}(\mathcal{U}_{2^{k}} (\hat{flow}_{k}),flow_{k}),
	\end{aligned}
\end{equation}
where $\mathcal{L}_{char}(\hat{x},x) = \sqrt{|| \hat{x} - x ||^{2} + {\varepsilon}^{2}}$ denotes the Charbonnier loss \cite{lai2017deep}, $\hat{x}$ and $x$ denotes the predicted results and their corresponding ground truth. $\varepsilon$ is empirically set to 10$^{-3}$,  $\mathcal{U}_s$ is the bilinear upsampling operation with scale
factor $s$, and $\alpha$ is a hyper-parameter, we set it to 10$^{-2}$.
For STNO-c, to enable the model to have perception capabilities at different scales and arbitrary intermediate time, we follow TMNet \cite{xu2021temporal} and perform generalization fine-tuning on Adobe \cite{SuDWSHW17} at multiple intermediate moments and scales. Specifically, we sampled the intermediate moment $t$ from \{1/8, 1/8, ..., 7/8\} and the upscaling factors from \{2.0, 2.2, ..., 3.8, 4.0 \} to train the STNO-c. %其他的实验设置和训练model-fix时类似。
The remaining experimental settings and training procedures are similar to those used for training the STNO-fix.
\begin{figure*}[!ht]
	\centering
	\includegraphics[width=15cm]{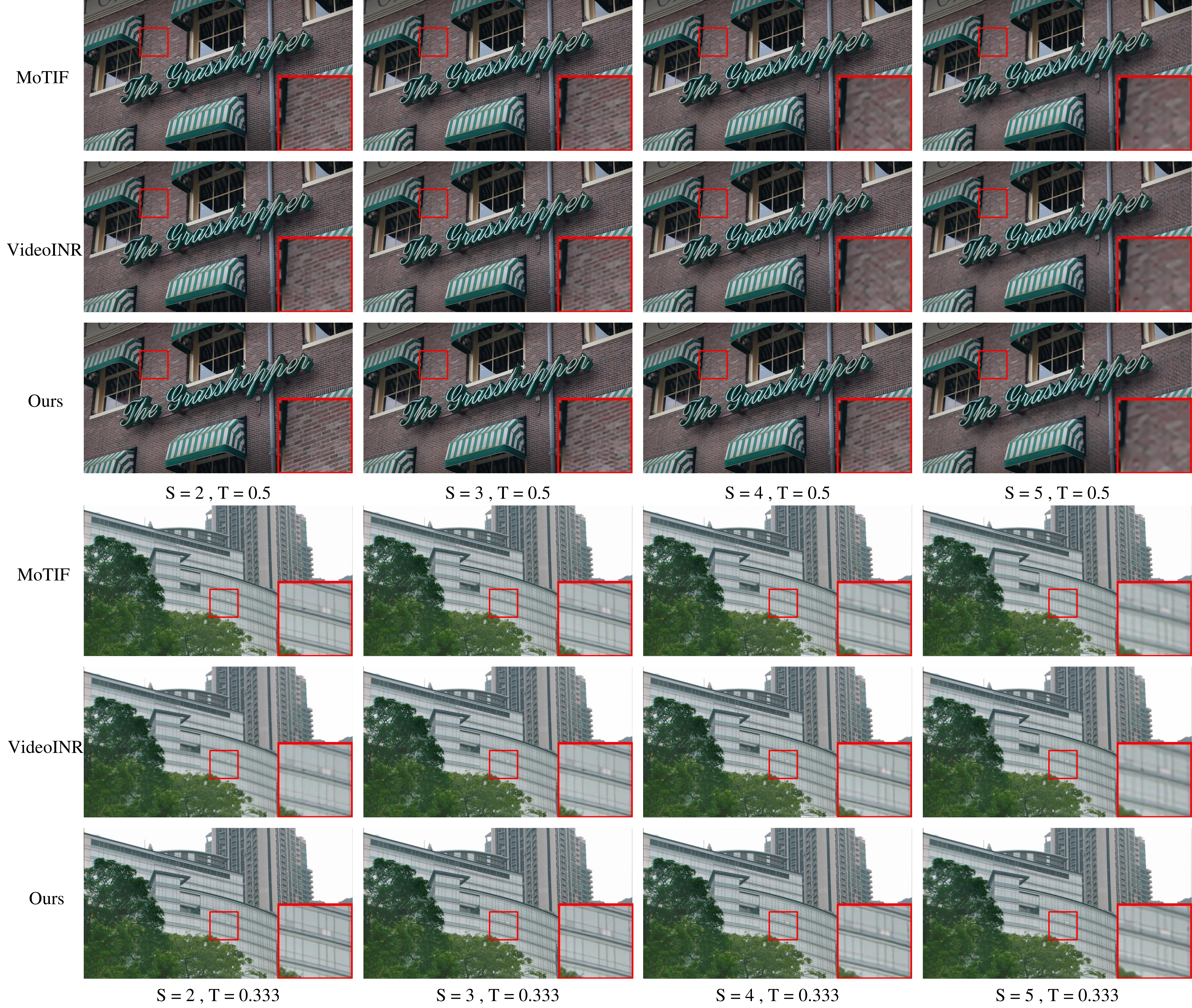}
	\caption{Comparison of continuous video spatiotemporal super-resolution performance on the SPMCS dataset, where S=A and T=B indicate spatial upsampling by a factor of A and intermediate temporal moments of B. } 
	%EDSC边缘处仍有伪影
	\label{fig6}
\end{figure*}
\subsection{Comparisons to State-of-the-Arts}
To evaluate the effectiveness of different super-resolution methods, we employ Peak Signal-to-Noise Ratio (PSNR) and structural similarity Index Measure (SSIM) as our performance metrics. Our comparative analysis encompasses both fixed and continuous Space-time video super-resolution methods. \\
\noindent \textbf{Comparison for Fixed ST-VSR.}
Since the majority of existing methods are limited to fixed spatial and temporal interpolation scales for space-time upsampling,
We first compare the results for space 4$\times$ and time 2$\times$ super-resolution with recent state-of-the-art one stage approaches \cite{2021Zooming,xu2021temporal,chen2022videoinr,chen2023motif} and two-stage approaches that sequentially apply VFI \cite{0Super,DAIN,EDSC,kong2022ifrnet,huang2020rife} and VSR \cite{2019EDVR,2020BasicVSR,chan2021basicvsr++}. 
From Table~\ref{tab1}, it can be observed that the proposed method outperforms all other methods on all metrics across all datasets. We summarize the key points as follows:
1) The performance of the proposed method improves progressively as the motion magnitude and texture complexity of the datasets increases. Due to the lower resolution of the Vimeo-90k-T dataset, the performance improvement of our method is not significant. Nevertheless, when evaluated on more challenging datasets like Adobe and GoPro, our method outperforms all existing solutions by a significant margin, achieving approximately 1 dB higher PSNR than the second-best method.
2) Thanks to the inherent resolution-independent characteristics of the neural operator, our model demonstrates strong generalization capabilities. Despite being trained only on Vimeo90k-T with lower resolution and smaller motion ranges, it achieves outstanding performance on both low-resolution and high-resolution data with varying motion ranges. In contrast, VideoINR\cite{chen2022videoinr} and MoTIF\cite{chen2023motif} exhibit significant performance degradation on Vimeo-90k-T, greatly limiting their adaptability to different scenarios.
3)  Some methods like VideoINR and MoTIF only consider information from two adjacent frames, ignoring valuable information from distant frames. However, most videos consist of more than just two frames.  Although we believe that this kind of evaluation approach is unreasonable, we still provide results using only two frames, which we refer to as "STNO-fix-two". It can be observed that even with only two frames, our results still outperform these approaches.
4) While achieving good flexibility, our method maintains the fastest speed and the smallest number of parameters. In comparison, VideoINR and MoTIF exhibit very slow inference speeds, as their complex additional motion encoding and feature aggregation components make their inference speed less than 1/10 of other approaches. 
In Fig.~\ref{fig4}, we present the restoration results of different methods on the Adobe\cite{SuDWSHW17} dataset. It can be observed that due to the camera shake, there is significant motion between adjacent frames. Both  deformable convolution kernels based \cite{2021Zooming,xu2021temporal} methods and optical based methods  \cite{chen2022videoinr,chen2023motif} fail to produce visually pleasing intermediate frames. In the first set of comparisons, all end-to-end space-time video super-resolution methods exhibit severe artifacts in the region between the two pillars, indicating that these methods fail to correctly capture the corresponding pixels in two adjacent frames.
\begin{figure*}[t]
	\centering
	\includegraphics[width=16cm]{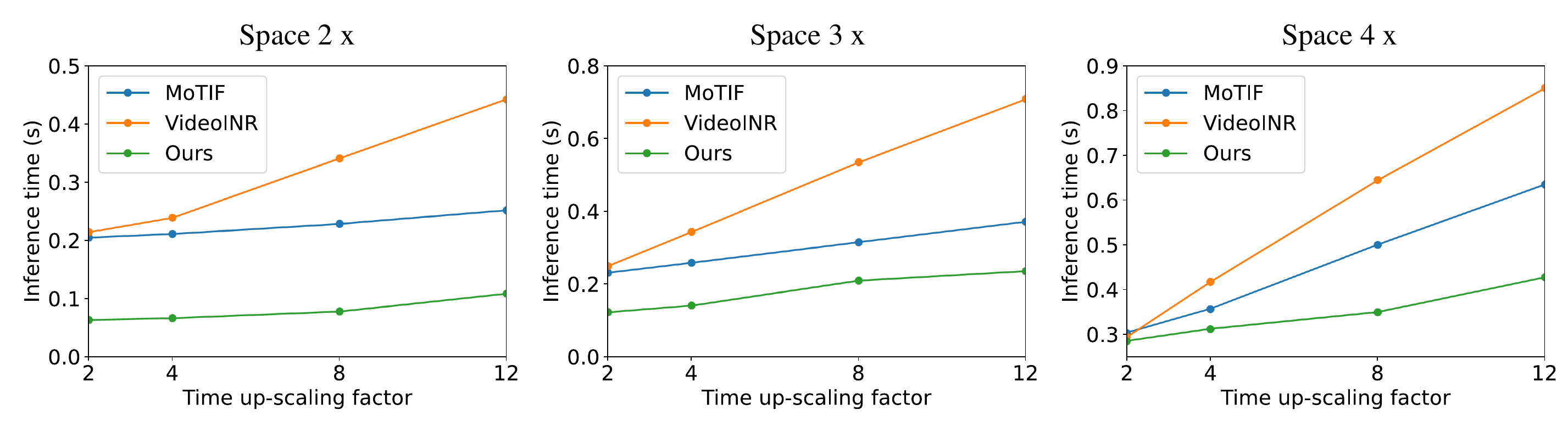}
	\caption{ Speed comparison of different continuous spatiotemporal super-resolution methods under various spatiotemporal upsampling rate combinations, our method outperforms the comparative algorithms in all tested cases. The input resolution is 128$\times$128, and the speed test was conducted on an NVIDIA 3090 GPU.
	}  \label{fig7}
\end{figure*}
In the second set of comparisons, results generated by VideoINR\cite{chen2022videoinr} and Zooming SloMo\cite{2021Zooming} exhibit noticeable ghosting effects between the bicyclist and the bicycle. Although TMNet\cite{xu2021temporal} and MoTIF\cite{chen2023motif} produce relatively better results, there is still a considerable degree of blurring. When we play multiple video frames in succession, it causes quality jitter between frames, which affects the overall viewing experience.  In contrast, some stronger two-stage methods (e.g., EDSC\cite{EDSC} + BasicVSR++\cite{chan2021basicvsr++}) and our proposed method yield significantly superior results. When compared to two-stage methods, our approach not only produces relatively sharper and closer-to-ground-truth results but also maintains a smaller parameter count and faster inference speed. We also present two sets of comparative results on GoPro\cite{DBLP:conf/cvpr/NahKL17}, where many scenes contain textual information. It can provide a good measure of the ability of different methods to maintain the continuity of texture patterns when synthesizing intermediate frames. As shown in Fig.~\ref{fig5}, all other competing methods (whether two-stage or one-stage STVSR schemes) fail to effectively restore text information. In contrast, our method has successfully restored text that is clearer and more complete, with sharper edges, making it easier to recognize and also closer to the ground truth. \\
\noindent \textbf{Comparison for Continuous ST-VSR.}
We then compare the results of continuous spatial-temporal upsampling. It has been noted in previous studies\cite{chen2022videoinr,chen2023motif} that a two-stage method involving sequential video frame interpolation and image super-resolution performs significantly poorer than a one-stage method.
In this context, our focus is on comparing against end-to-end spatio-temporal super-resolution methods. 
Specifically, we compared different combinations of intermediate moments and resolutions. The experimental results are provided in Table~\ref{tab2}. 
It can be observed that our method significantly outperforms VideoINR and MoTIF, both for in-distribution and out-distribution results. In certain space-time combinations (e.g., T  $\times$ 2, S $\times$ 2) the performance improvement reaches up to 3 dB. 
In Fig.~\ref{fig6}, we present the visual comparison results for different intermediate time steps and spatial resampling rate combinations. It can be seen that the proposed method exhibits significantly superior subjective visual quality, consistent with objective metrics. For instance, in the first set of images with S=4 and T=0.5, both VideoINR and MoTIF exhibit severe aliasing artifacts. In contrast, our method achieves superior inter-frame information alignment and restores more reliable texture details by leveraging abundant multi-frame information.
\subsection{Computation Complexity Analysis}
\label{complexity}
% 这一部分我们分析
\begin{table}[!t]
	\centering
	\caption{
		Evaluation for computational complexity (GFLOPs) on various datasets,  we perform space 4 $\times$ and time 2 $\times$ upsampling. To accommodate certain methods that require the input image resolution to be a multiple of 8, we perform padding on the input image if needed.
	}
	\resizebox{0.99\linewidth}{!}{	
		\begin{tabular}{c|cccc}
			\hline
			Dataset    & Vid             & Vimeo           & Adobe \& GoPro   & SPMCS            \\ \hline
			Input Size & 144 $\times$ 184 & 64 $\times$ 112 & 184 $\times$ 320 & 128 $\times$ 240 \\ \hline
			VideoINR   & 924.85          & 250.20          & 2055.21          & 1072.29          \\
			MoTIF      & 1125.52         & 304.47          & 2501.32          & 1305.12          \\
			STNO (Ours)       & 543.03          & 146.91          & 1206.74          & 629.60           \\ \hline
		\end{tabular}
	}\label{tab4}
\end{table}
We carefully consider reducing the computational complexity in model design. Firstly, the Galerkin-type attention exhibits a linear complexity relationship with the input resolution. In contrast, many optical flow estimation networks \cite{2016FlowNet,DBLP:conf/eccv/TeedD20,2017PWC} introduce correlation layers with quadratic complexity for motion estimation. Secondly, we have not designed a separate motion estimation module; instead, motion estimation and motion compensation are performed in a coupled manner. This compact design not only reduces the number of parameters but also enhances computational efficiency. On the contrary, some methods that introduce independent optical flow estimation networks often require additional post-processing modules (such as Gridnet in SoftSplat \cite{2020Softmax} or Unet in \cite{zhang2023extracting}) to reduce artifacts. 
Quantitative experiments also validate this point. In Fig.~\ref{fig7}, we provide a speed comparison of various methods under different spatiotemporal super-resolution factors. It can be observed that the proposed approach achieves significantly faster inference speeds compared to VideoINR and MoTIF under all conditions. Additionally, Table~\ref{tab4} presents the GFLOPs of different methods on various test datasets. Our approach maintains the lowest floating point operations across different input resolutions, approximately half of the comparison methods. These experimental results demonstrate that the proposed approach not only achieves excellent performance but also significantly reduces computational complexity and inference time.
\begin{figure*}[h]
	\centering
	\includegraphics[width=16.0cm]{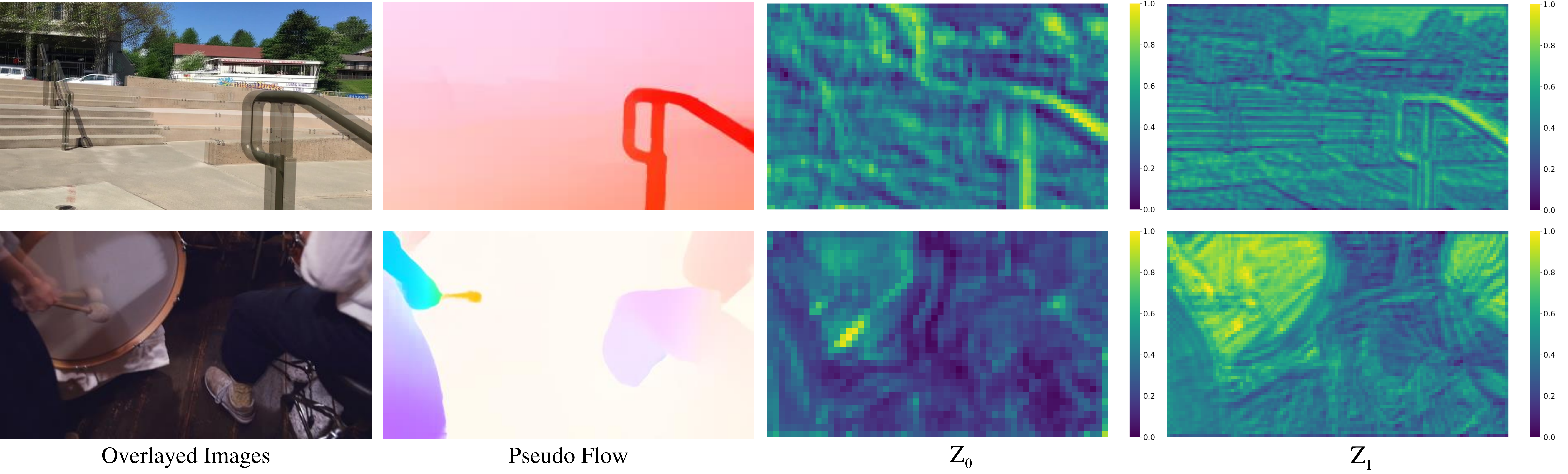}
	\caption{ Visual of the Galerkin type attention. 
		From left to right, we display the overlapping consecutive frames, the pseudo-optical flow $flow_{0 \rightarrow 1}$ and the normalized output ($z$ in Eq.~\ref{eq7}) of the Galerkin-type attention module. $Z_{i}$
		represents the 
		$i^{th}$ iterative layer.
	} \label{atten}
\end{figure*}
\begin{figure*}[h]
	\centering
	\includegraphics[width=16.0cm]{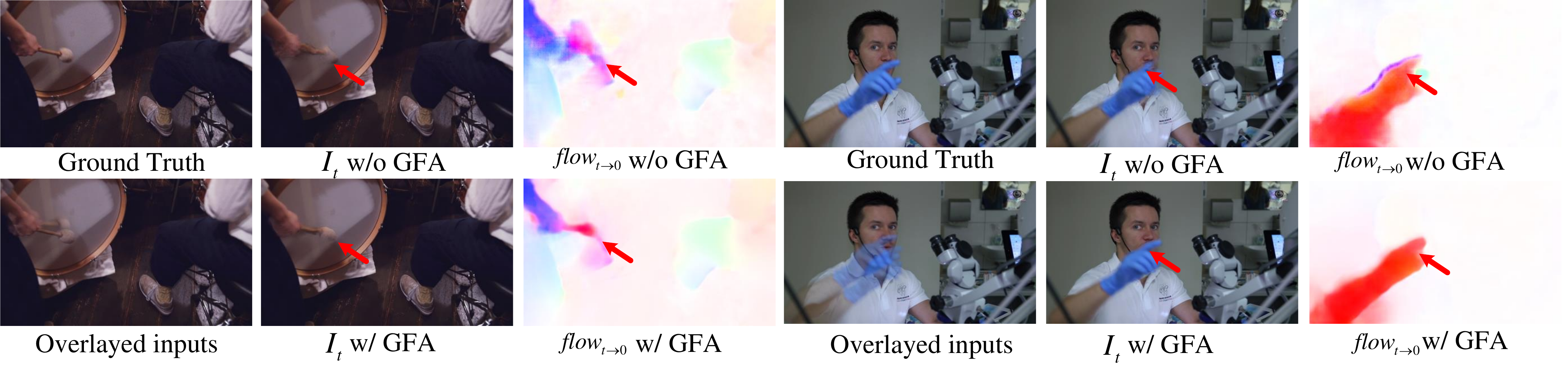}
	\caption{ Visual comparison of w/wo Global Feature Aggregation (GFA) module. We also provide visualization results of the corresponding estimated motion flow.
	} \label{flow_cp}
\end{figure*}
\subsection{Visualization of Galerkin-type kernel Function}
\label{kernel}
To better understand the working mechanism of Galerkin-type attention, we visualize the output of this module.  As shown in Fig.~\ref{atten}, we present two consecutive frames and calculate the optical flow pseudo-label between the two frames using a state-of-the-art optical flow estimation model\cite{JiangCLLH21}. We also generate the normalized heatmap of attention ($z$ in Eq~\ref{eq7}). It can be observed that this module pays more attention to the regions with motion between the two frames. Additionally, $z_{0}$ focuses more on areas with significant motion, while $z_{1}$ emphasizes relatively subtle motion. This behavior arises from the iterative neural operator architecture, which allows the $z_{i+1}$ motion estimation to effectively capture the residual motion from the $z_{i}$ estimation. 
\subsection{Ablation Study}
\begin{table}[t]
	\centering
	\caption{Ablation study of global texture aggregation and coupled MEMC. Quality
		metrics: PSNR(dB)/SSIM }
	\resizebox{0.93\linewidth}{!}{	
		\begin{tabular}{ccc|ccc}
			\hline
			Case &TFA & MFA & Vid          & Adobe        & GoPro        \\ \hline
			1 &\ding{55}   & \ding{55}   & 26.27/0.7977 & 30.02/0.8678 & 30.41/0.8769 \\ \hline
			2 &\ding{55}   & $\checkmark$   & 26.37/0.7988 & 30.01/0.8659 & 30.44/0.8762 \\ \hline
			3 &$\checkmark$   & \ding{55}   & 26.50/0.8052 & 31.01/0.8914 & 31.39/0.8954 \\ \hline
			4 &$\checkmark$   & $\checkmark$   & 26.72/0.8138 & 31.85/0.9073 & 32.06/0.9060 \\ \hline
		\end{tabular}
	}\label{tab3}
\end{table}
\noindent \textbf{Global Feature Aggregation.}
As the most crucial part of this work, we craft a Galerkin-type attention including both texture feature aggregation (TFA) and motion feature aggregation (MFA) as the kernel function of the neural operator for global feature aggregation (GFA). 
Benefiting from the linear complexity presented by global feature aggregation for the input resolution, we do not need to perform any patch partition operations on the input features, which efficiently achieves a global receptive field.
To validate the effectiveness of this module, we conduct an ablation study by omitting this mechanism and solely relying on optical flow warping for alignment purposes.  The experimental results are presented in Table~\ref{tab3}. It can be observed that removing the global MEMC (Case 1) results in severe performance degradation across all datasets. Specifically, the performance improvement brought by global feature aggregation is  related to the characteristics of the dataset. Since the four sequences in the Vid dataset have relatively small motion ranges, the GFA module provides a modest improvement of around 0.45dB in terms of PSNR. However, for datasets with larger and more challenging motion ranges like Adobe and GoPro, the performance improvement is significant, reaching around 1.8 dB.
 In Fig.~\ref{flow_cp}, we present two sets of typical visual results, each consisting of interpolated intermediate frames and their corresponding motion flow. It can be observed that when the attention mechanism is absent,  it leads to a significant decrease in performance.  
 In Table~\ref{optical}, we first utilize a state-of-the-art pre-trained optical flow extractor \cite{JiangCLLH21} to compute the optical flow for the Vimeo-90K-T dataset, which serves as pseudo-labels. Using these pseudo-labels, we then evaluate the model w/wo global feature aggregation based on the End Point Error (EPE). As shown, model with GFA significantly improves the accuracy of the optical flow estimation. Furthermore, the improvement becomes more apparent as the magnitude of motion increases.
\begin{table}[!t]
	\centering
	\caption{The impact of w/wo GFA on the quality of optical flow. We report the End Point Error (EPE) of predicted optical flows and their corresponding pseudo-labels.}
	\resizebox{0.99\linewidth}{!}{	
		\begin{tabular}{c|cccc}
			\hline
			& Vimeo-Slow  & Vimeo-Medium    & Vimeo-Fast         & Vimeo-Avg\\ \hline
			w/o GFA   & 0.43        & 0.96           &2.87                 & 1.15    \\ \hline
			w/ GFA    & 0.35        & 0.84           & 2.59                & 1.01    \\ \hline
			Difference         &0.08         &0.12            & 0.28                & 0.14      \\ \hline
		\end{tabular}
	}\label{optical}
\end{table} \\
\noindent \textbf{Coupled MEMC.} We perform progressive coupling estimation of texture and motion information using a Galerkin-type kernel, rather than designing a separate optical flow estimation module. The motivation behind coupled MEMC is as follows: at each spatial scale, more reliable texture information provides a smoother input for motion estimation, while more accurate motion information, in turn, helps synthesize more plausible intermediate frame textures.
To validate the effectiveness of this mechanism, we retained only one of the two modules, TFA or MFA. (corresponding to Case 2 and Case 3 in Table~\ref{tab3}). The results demonstrate that the absence of either module has a significant impact on performance.
Another advantage of the coupled motion estimation is that we no longer need separate optical flow estimation modules (e.g., Spynet \cite{ranjan2017optical} for BasicVSR \cite{2020BasicVSR}, Raft \cite{DBLP:conf/eccv/TeedD20} for MoTIF \cite{chen2023motif}). This reduces the complexity of the model and allows motion information and texture information to promote each other, resulting in accurate alignment and more precise temporal interpolation.
\section{Conclusions and Future Work}
\label{conclu}
In this paper, we present a novel approach to address the space-time video super-resolution (ST-VSR) task by formulating it as a neural operator learning problem. By conceptualizing the problem as a mapping between two function spaces, we aim to extract fine-grained spatiotemporal information from coarse-grained intra-frame features. 
Specifically, we adopt a Galerkin-type kernel integral in our neural operator, which enhances motion estimation's precision and efficiency due to its global receptive field and linear complexity.  
 Further, our model consolidates motion information for alignment and interpolation, eliminating redundant motion estimation and compensation calculations and enhancing overall efficiency. 
Extensive experiments demonstrate that our method outperforms state-of-the-art techniques in both fixed-size and continuous space-time video super-resolution tasks with faster speed and reduced parameter count.

Despite the significant improvements in performance and efficiency, there are still some limitations. Firstly, we focus on efficient and accurate motion estimation and compensation (MEMC), without explicitly considering the utilization of intra-frame information. Furthermore, in our approach, the utilization of neural operator primarily focuses on addressing the MEMC problem. This is attributed to the inherent complexity of capturing real-world motion, which poses a challenge in establishing an end-to-end neural operator framework that directly achieves both temporal and spatial super-resolution. We leave these issues for future work.

\bibliographystyle{IEEEtran}
\bibliography{11_references}

\end{document}